\documentclass{article}

\usepackage{times}
\usepackage{graphicx} 
\graphicspath{ {images/} }
\usepackage{subfigure} 

\usepackage{algorithm}
\usepackage{algorithmic}

\usepackage[utf8]{inputenc} 
\usepackage{hyperref}       
\usepackage{url}            
\usepackage{booktabs}       
\usepackage{amsfonts}       
\usepackage{nicefrac}       
\usepackage{microtype}      

\usepackage{amsmath}
\usepackage{amsthm}
\usepackage[english]{babel}

\newtheorem{theorem}{Theorem}
\newtheorem{lemma}[theorem]{Lemma}

\newcommand{\norm}[1]{\left\lVert#1\right\rVert}

\newcommand{\scalefactor}{0.26}
\newcommand{\proxyscalefactor}{0.36}

\usepackage{hyperref}

\usepackage[accepted]{icml2017}

\icmltitlerunning{Practical Learning of Predictive State Representations}

\begin{document} 

\twocolumn[
\icmltitle{Practical Learning of Predictive State Representations}



\icmlsetsymbol{equal}{*}

\begin{icmlauthorlist}
\icmlauthor{Carlton Downey}{cmu}
\icmlauthor{Ahmed Hefny}{cmu}
\icmlauthor{Geoffrey Gordon}{cmu}
\end{icmlauthorlist}

\icmlaffiliation{cmu}{Carnegie Mellon University, Pittsburgh, PA, USA}

\icmlcorrespondingauthor{Carlton Downey}{cmdowney@cs.cmu.edu}
\icmlcorrespondingauthor{Ahmed Hefny}{ahefny@cs.cmu.edu}
\icmlcorrespondingauthor{Geoffrey Gordon}{ggordon@cs.cmu.edu}

\icmlkeywords{}

\vskip 0.3in
]



\printAffiliationsAndNotice{} 

\begin{abstract}
Over the past decade there has been considerable interest in spectral algorithms for learning Predictive State Representations (PSRs). Spectral algorithms have appealing theoretical guarantees; however, the resulting models do not always perform well on inference tasks in practice. One reason for this behavior is the mismatch between the intended task (accurate filtering or prediction) and the loss function being optimized by the algorithm (estimation error in model parameters).

A natural idea is to improve performance by refining PSRs using an algorithm such as EM. Unfortunately it is not obvious how to apply apply an EM style algorithm in the context of PSRs as the Log Likelihood is not well defined for all PSRs. We show that it is possible to overcome this problem using ideas from Predictive State Inference Machines \cite{sun2016}.

We combine spectral algorithms for PSRs as a consistent and efficient initialization with PSIM-style updates to refine the resulting model parameters. By combining these two ideas we develop Inference Gradients, a simple, fast, and robust method for practical learning of PSRs. Inference Gradients performs gradient descent in the PSR parameter space to optimize an inference-based loss function like PSIM. Because Inference Gradients uses a spectral initialization we get the same consistency benefits as PSRs. We show that Inference Gradients outperforms both PSRs and PSIMs on real and synthetic data sets.

\end{abstract}

\section{Introduction}

Predictive state representations (PSRs) \cite{Littman01predictiverepresentations} are a class of models for filtering, prediction, and simulation of discrete time dynamical systems. PSRs provide a compact representation of a dynamical system by representing state as a set of predictions of features of future observations. This representation is known as a \emph{predictive state}, in contrast to the \emph{latent state} present in models such as Hidden Markov Models (HMMs). 

PSRs are an important class of models because, due to the (noisy but direct) observability of their parameters, there exist spectral algorithms for learning PSR model parameters which are statistically consistent, computationally efficient, and globally optimal. In contrast, competing techniques based on the likelihood function (such as Expectation Maximization (EM)) are often slow to converge \cite{wu1983} and are highly susceptible to local optima.  


Despite the appealing theoretical properties of spectral algorithms for PSRs these models have seen limited use in practice due to underwhelming experimental performance on many problems. In an effort to overcome this issue several authors have proposed using an EM style algorithm to post process/fine tune the model parameters of a PSR learned using a spectral algorithm \cite{jks2016,shaban:2015}. Unfortunately EM algorithms are based on the log likelihood function, which is not well defined for all PSRs. Specifically, filtering and prediction in PSRs learned using a spectral algorithm can produce states which are unbounded in size, or which correspond to negative probabilities.


We show that this problem can be solved by considering an alternative objective function which is not based on the log likelihood. We present a simple, computationally efficient algorithm which allows us to fine tune the model parameters of an arbitrary PSR and results in large improvements in experimental performance.



Sun et al. \cite{sun2016} introduce the idea of Predictive State Inference Machines (PSIMs) which directly train models to perform inference. They show that models trained in this way can outperform both latent state models and models trained via spectral techniques. Furthermore they establish asymptotic and finite sample bounds on the filtering performance of the resulting models. However the form of these results implies that the finite sample performance of their algorithm may depend heavily on how it is initialized. Furthermore the results assume that the training of the inference model can be performed perfectly, which may not hold for the non-linear filtering models that can emerge in discrete systems. In this setting, having a good initialization becomes even more important.

We combine spectral algorithms for PSRs as a consistent and efficient initialization with ideas from PSIMs to refine the model parameters in order to achieve good predictive performance. By combining  these two ideas we get Inference Gradients, a simple, fast, and robust method for practical learning of PSRs. Inference Gradients allows us to perform gradient descent in the PSR parameter space to optimize an inference based loss function without being forced to define a proxy to the log liklihood. We show that Inference Gradients outperforms both PSRs and PSIMs on two real and synthetic data sets.

\section{Background}

\subsection{Dynamical Systems}
A dynamical system is a stochastic process (i.e., a distribution over sequences of observations) such that, at any time, the distribution of future observations is fully determined by a vector $s_t$ called the latent state. Note that the distribution of $s_t$ depends only on history. The process is specified by three distributions: the initial state distribution $P(s_1)$, the state transition distribution $P(s_{t+1} \mid s_t)$, and the observation distribution $P(o_t \mid  s_t)$. Given a dynamical system, one of the fundamental tasks is to perform inference, where we predict future observations given a history of observations. Typically this is accomplished by maintaining a distribution or belief over latent states $b_{t\mid t-1} = P(s_t \mid  o_{1:t-1})$, where $o_{1:t-1}$ denotes the first $t - 1$ observations. $b_{t\mid t-1}$ represents both our knowledge and our uncertainty about the true state of the system. 

Two core inference tasks are filtering and prediction. In filtering, given the current belief $b_t = b_{t\mid t-1}$ and a new observation $o_t$, we calculate an updated belief $b_{t+1} = b_{t+1\mid t}$ that incorporates $o_t$. In prediction, we project our belief into the future: given a belief $b_{t\mid t-1}$ we estimate $b_{t+k\mid t-1} = P(s_{t+k} \mid  o_{1:t-1})$ for some $k > 0$ (without incorporating any intervening observations).

\subsection{Predictive State Representations (PSRs)}
The classical approach for modelling a dynamical system is to explicitly estimate the initial, transition, and observation distributions. Estimates of these distributions can be used compute and update a belief over latent states, which in turn allows us to make predictions of future observations. PSRs take an alternative approach: instead of maintaining a belief $b_t$ over latent states $s_t$, they maintain a \emph{predictive state} represented by the expected value of a sufficient statistic of future observations \cite{Jaeger99observableoperator, Littman01predictiverepresentations,Singh:2004,boots:2011,boots:2009:closing,hefny:2015}. In this work we will use the PSR formulation of \cite{hefny:2015,Venkatraman:2016:online}.

In more detail, we define a predictive state $q_t = q_{t\mid t-1} = E[\psi_t \mid  o_{1:t-1}]$, where $\psi_t = \psi(o_{t:t+k-1})$ is a vector of features of future observations. The features are chosen such that $q_t$ determines the distribution of future observations $P(o_{t:t+k-1} \mid  o_{1:t-1})$.\footnote{For convenience we assume that the system is $k$-observable: that is, the distribution of all future observations is determined by the distribution of the next $k$ observations. (Note: not by the next $k$ observations themselves.) 
At the cost of additional notation, this restriction could easily be lifted.}
Filtering then becomes the process of mapping a predictive state $q_t$ to $q_{t+1}$ conditioned on $o_t$, while prediction maps a predictive state $q_t = q_{t\mid t-1}$ to $q_{t+k\mid t-1} = E[\psi_{t+k} \mid  o_{1:t-1}]$ without intervening observations. 

PSRs are based on the idea that it is sufficient to maintain $q_t$ in order to make predictions about future observations conditioned on the history. The key benefit offered by PSRs is that by using an observable representation of state we can develop efficient globally convergent algorithms \cite{Singh:2004}.



\subsection{Spectral Learning of PSRs}
\label{sec:hefny}

The classical approach to learning a dynamical system is to optimize the model parameters by maximizing the likelihood of the observed data. Unfortunately the likelihood function is often highly non-convex, leading to local optima and sub-optimal model parameters. Spectral algorithms offer an alternate approach to learning: they use the method of moments to set up a system of equations that can be solved in closed form to recover estimates of the desired parameters. In this process, they typically factorize a matrix or tensor of observed moments—hence the name ``spectral''.

Hefny et al. \cite{hefny:2015} show that spectral learning of dynamical systems can be formulated as solving a sequence of regression problems. We follow this approach, referred to as \emph{two-stage regression} or \textbf{2SR}, in our work. Under this framework, spectral learning of PSRs corresponds to learning a set of regression models. In the case of a discrete observation space\footnote{While we focus on the discrete setting for ease of exposition, the main idea of this work can be extended to continuous settings.}, it can be shown that two stage regression with linear models is equivalent to learning an ordinary PSR, that is, an initial state $q_1$, a normalizer $b_{\infty}$, and set of linear operators $B_i$ for $i \in \{1,...,k\}$ such that:

\begin{align*}
\label{eqn:psr_update}
    q_{t+1} =  \frac{B_{o_t}q_t}{b_{\infty}^T B_{o_t}q_t}
\end{align*}

Given a set of training examples ($h_t,o_t,\psi_t,\psi_{t+1}$) for $1 \leq t \leq T$, we can estimate these parameters as follows:

\begin{align}
    q_1 & = \frac{1}{T}\sum_{t=1}^T \psi_t \\
    B_i & = \left(\sum_{t=1}^T \psi_{t} h_t^\top \right)^+ \left(\sum_{t=1}^T 1(o_t = i) \psi_{t+1} h_t^\top \right) \\
    b_\infty^T & = \left( \sum_{t=1}^T h_t^\top \right) \left(\sum_{t=1}^T \psi_{t} h_t^\top \right)^+
\end{align}

Note that while each $B_i$ is linear, the state update is \emph{non-linear} due to the normalization term. 

\subsection{Predictive State Inference Machines (PSIMs)}
PSIMs \cite{sun2016} constitute an inference method for dynamical systems that
combines (1) the idea of a predictive state from PSRs (where the state is a prediction of future statistics)
and (2) inference machines \cite{langford:2009:learning,ross:2011} where, instead of learning generative model parameters that are then fed into a fixed inference function, we directly learn the inference function.

PSIMs directly learn an inference function by minimizing predictive loss on the training set. In contrast, a PSR specifies a data \emph{generation} model, and the goal of the spectral algorithm
is to identify the parameters of this model by matching training statistics. These parameters indirectly determine the inference function.

More formally, the goal is to learn a function $f$ that can deterministically pass the predictive states forward in time conditioned on the latest observation -- that is, $\hat{q}_{t+1} = f(\hat{q}_t, o_t)$ such that the likelihood of the observations $o_t$ generated from the sequence of predictive states is maximized.  

As this is a sequential prediction problem over the predictive states,we use DAgger \cite{ross:2011} to optimize the inference machine. The choice of learner for $f$ can be any no-regret regression or classification algorithm.

We note that PSIM updates can be implemented as gradient descent on the loss function $L(f) = \frac{1}{T}\sum_{t=1}^T\norm{\hat{q}_t - q_t}^2_2$ where $\hat{q}_t$ is the state estimate at time $t$ produced by filtering, while $q_t$ is the true state at time $t$. This procedure is outlined in Algorithm \ref{alg:psim}:

\begin{algorithm}[tb]
   \caption{PSIM}
   \label{alg:psim}
\begin{algorithmic}
   \STATE {\bfseries Input:} sequence of observations $o_{1:T}$, learning rate $\alpha \in \mathbb{R}^+$, number of iterations $n \in \mathbb{N}^+$.
   \STATE
   \STATE $f \leftarrow$ An arbitrary hypothesis.
   \FOR{$i=1$ {\bfseries to} $n$}
   \STATE $\hat{q}_1 \leftarrow q_1$
   \FOR{$t=1$ {\bfseries to} $T$}
   \STATE $\hat{q}_{t+1} \leftarrow f(\hat{q}_t, o_t)$
   \STATE $f \leftarrow f - \alpha \Delta L(f)$
   \ENDFOR
   \ENDFOR
\end{algorithmic}
\end{algorithm}

\section{Related Work}

The classic approach for refining a dynamical system model is the EM algorithm. EM iteratively adjusts the model parameters to locally optimize the log likelihood. Unfortunately inference in PSRs can produce states which are unbounded in size, or which correspond to negative probabilities. This means that the log likelihood is not well defined for arbitrary PSR parameters. Several authors attempt to solve this problem in a variety of ways. 

Jiang et al. \cite{jks2016} propose a gradient descent algorithm for improving the performance of PSRs where they optimize a proxy to the log likelihood. Given a PSR $\mathcal{B} = (b_1, b_{\infty}, \{B_i\}_{i=1}^\infty)$ and a sequence of observations $o = o_1,...,o_n$, the negative log likelihood is:
\begin{align}\label{eqn:pnll}
\ell(\mathcal{B};o) = -\log(P_\mathcal{B}(o)) = -\log(b_{\infty}B_n...B_1 b_1)
\end{align}
Because the log likelihood is not well defined for an arbitrary PSR, Jiang et al. choose to optimize a related loss function which rectifies and re-normalizes each predicted observation probability distribution:
$$
\ell(\mathcal{B};o) = -\log\left(\frac{ \vert P_\mathcal{B}(o) \rvert}{\sum_{y \in O^{\lVert o \rVert}}\lvert P_\mathcal{B}(y) \rvert}\right)
$$
where $O^{\lVert o \rVert}$ is the space of all observation sequences with the same length as $o$. This yields the gradient
\begin{align*}
\frac{d}{dB_i}\ell(\mathcal{B};o) &= \frac{d}{dB_i}\log\left(\sum_{y \in O^{\lVert o \rVert}}\lVert P_\mathcal{B}(y)\rVert \right)\\ 
&- \frac{d}{dB_i}\log \lVert P_\mathcal{B}(o)\rVert
\end{align*}
This expression is analytically intractable, so they propose a stochastic gradient descent procedure where they approximate this expression using contrastive divergence. They show that this approach can be used to significantly improve the performance of a PSR initialized via spectral techniques. This approach allows for gradient descent on a surrogate to the log loss, however the resulting algorithm has a complex update rule, and can be slow in practice.

Shaban et al. \cite{shaban:2015} propose a two pass algorithm for learning a dynamical system model, where they first learn a PSR using a spectral algorithm, then subsequently convert the PSR into a valid HMM using an exterior point method. Their approach is different from ours as they produce an HMM as the final model rather than a PSR. Additionally, their algorithm focuses on model parameter optimization rather than optimal performance on inference tasks.

Sun et al. \cite{sun2016} (see background) attempt to learn an inference model directly using the observable features of PSRs, but without the PSR model. This approach means that they are no longer learning a model class which includes HMMs. Furthermore their algorithm requires a good initialization in order to perform well.

These methods offer significant improvements over PSRs trained using two-stage regression on many problems, however they also come with disadvantages. We would like an algorithm with PSIM-like behavior, but (1) on a model class such as PSRs and (2) which includes HMMs and a good initialization.

\section{Inference Gradients}
\label{sec:ig}

One reason that two-stage regression for PSRs can perform poorly on finite data sets is the mismatch between the intended task (accurate filtering or prediction) and the loss function being optimized by the algorithm (estimation error in model parameters). In the realizable setting, under appropriate regularity conditions, spectral algorithms asymptotically converge to the true model parameters. Unfortunately these results tell us little about how models learned from finite data will perform on filtering or prediction tasks in practice. Indeed, PSRs learned using a spectral algorithm can produce poor predictions when applied recursively to the original training data, even if they are close to the true model parameters \cite{sun2016}! Rather than a guarantee that the learned model parameters will be close to the true model parameters, we would prefer a guarantee that the learned model will perform well on inference tasks.

Additional insight into this problem can be gained by examining how we learn a model using two stage regression, and what happens when that model is used to perform filtering. Let $(q_1, \{B_i\}_{i=1}^k, b_{\infty})$ be a discrete PSR learned via two stage regression as described in Section \ref{sec:hefny}. Let $o_1,...,o_T$ be a sequence of observations and $q_1,...,q_T$ the corresponding sequence of true states. Let $\hat{q}_t$ be the estimated belief over states at time $t$ produced by filtering. Given state $q_t$ we apply the PSR update $\hat{q}_{t+1} = \frac{B_{o_t}\hat{q}_t}{b_{\infty}^T B_{o_t}\hat{q}_t}$ to produce $\hat{q}_{t+1}$.

It can be seen from Equation (2) that each $B_i$ is the solution
to a regression problem from expected features of the future at time $t$ to expected features of the future at time $t+1$. In other words we learn a model which makes good one step inference predictions for all states $q_t$ encountered in the training set. Unfortunately, using (2) to perform inference results in states not present in the training set due to model estimation error. This causes poor subsequent predictions, as our model was not optimized to make good predictions on states outside of the training set.

We would like to modify the learned PSR so that it makes good predictions on states encountered during inference, not just the states encountered when solving the two-stage regression. 

To solve this problem, we turn to PSIM. The PSIM framework specifically trains a model to make good predictions on states encountered during inference. Essentially, PSIM uses an initial model to perform filtering on a sequence of observations to produce a sequence of states, then updates the model so that the states produced during inference are closer (in expectation) to the true states.

One issue with PSIM is that the training data for PSIM is generated from the initial model. If the initial model is poor, the resulting training data will also be poor, and updating the model based on this data is likely to be of little benefit. Adapting PSIM to PSRs allows us to generate a good PSIM initialization using two-stage regression. This offers an alternative viewpoint: that two-stage regression applied to PSR allows us to fully unlock the potential of PSIMs by providing a consistent way to initialize them.

A second issue is that it is not obvious how to apply PSIM to PSRs. In PSRs we use a weird functional form consisting of a linear update followed by a normalization step. This functional form does not fit nicely within the PSIM framework. We show that by using the gradient descent fomulation of PSIM it is possible to perform PSIM-style updates on an arbitrary PSR.

We apply PSIM-style updates to a PSR initialized using 2-stage regression. Our proposed updates correspond to gradient descent on the loss function $L(\{B_i\}_{i=1}^k) = \sum_t \frac{1}{2}\norm{q_{t+1} - \hat{q}_{t+1}}_2^2$. Since $q_t$ is not observed we replace it with the observed features of future observations $\psi_t$ as an unbiased estimate of $q_t$ giving us the loss function.
\begin{align}
\label{eqn:loss_function}
\widehat{L}(\{B_i\}_{i=1}^k) = \frac{1}{T}\sum_{t=1}^T \frac{1}{2}\norm{\psi_{t+1} - \hat{q}_{t+1}}_2^2
\end{align}
The gradient of this function can be found in Lemma \ref{lemma:1step}. The proof of Lemma \ref{lemma:1step} can be found in Appendix \ref{apdx:1step}.

\begin{lemma}
\label{lemma:1step}
\begin{align*}
\frac{\partial}{\partial B_{o_t}} \norm{\psi_{t+1} - \hat{q}_{t+1}}_2^2
&=\frac{\partial}{\partial B_{o_t}} \norm{\psi_{t+1} - \frac{B_{o_t}\hat{q}_t}{b_{\infty}^T B_{o_t}\hat{q}_t}}_2^2\\
&= \left(\frac{b_{\infty}\hat{q}_{t+1}^T - I}{b_{\infty}^T B_{o_t}\hat{q}_t}\right)(\psi_{t+1} - \hat{q}_{t+1})\hat{q}_t^T
\end{align*}
\end{lemma}

This gradient combined with 2-stage regression and PSIM style updates gives us the following simple algorithm for learning PSRs:

\begin{algorithm}[tb]
   \caption{Learning a PSR with One-Step Inference Gradients}
   \label{alg:psr}
\begin{algorithmic}
   \STATE {\bfseries Input:} sequence of observations $o_{1:T}$,  corresponding set of features of future observations $\psi_{1:T}$, learning rate  $\alpha \in \mathbb{R}^+$, number of iterations $n \in \mathbb{N}^+$.
   \STATE
   \STATE $(q_1, \{B_i\}_{i=1}^k, b_{\infty}) \leftarrow$ \textbf{2S-Regression}($o_{1:T}$, $\psi_{1:T}$)
   \FOR{$i=1$ {\bfseries to} $n$}
   \STATE $\hat{q}_1 \leftarrow q_1$
   \FOR{$t=1$ {\bfseries to} $T$}
   \STATE $\hat{q}_{t+1} \leftarrow \frac{B_{o_t}\hat{q}_t}{b_{\infty}^T B_{o_t}\hat{q}_t}$
   \STATE $B_{o_t} \leftarrow B_{o_t} - \alpha \left(\frac{b_{\infty}\hat{q}_{t+1}^T - I}{b_{\infty}^T B_{o_t}\hat{q}_t}\right)(\psi_{t+1} - \hat{q}_{t+1})\hat{q}_t^T$
   \ENDFOR
   \ENDFOR
\end{algorithmic}
\end{algorithm}

\section{Multi-step Inference Gradients}
In step $t$ of Algorithm \ref{alg:psr} we take a gradient step such that $\hat{q}_{t+1}$  is closer to the true state $\psi_{t+1}$. However it is important to note that changing $B_{o_t}$ will also change future predicted sates $\hat{q}_{t+2},\hat{q}_{t+3},\mbox{etc}$. In fact making $\hat{q}_{t+1}$ closer to $\psi_{t+1}$ may actually cause future predicted states to be worse. This effect is not taken into account in PSIM, where the objective function being optimized is the mean squared loss of all one-step updates.

This discussion suggests that we should be optimizing each $B_i$ with respect to its long term effect on inference predictions. With this in mind we extend the results from the previous section from one-step Inference Gradients to $h$-step Inference Gradients. In an $h$-step inference gradient we take into account the effect of changing the observable operator at the current observation on the error in the $h$th future observation. 

\begin{lemma}
\label{lemma:multi}
\begin{align*}
&\frac{\partial}{\partial B_{o_t}} \frac{1}{2}\norm{ \psi_{t+h} - \hat{q}_{t+h} }_2^2\\ 
&=B_{o_{t+1:t+h}}^T\left(\frac{b_{\infty}\hat{q}_{t+h}^T - I}{b_{\infty}^T B_{o_{t:t+h}}\hat{q}_t}\right)(\psi_{t+h} - \hat{q}_{t+h})\hat{q}_t^T
\end{align*}
where $B_{o_{t:t+h}}$ is defined as $B_{o_{t:t+h}} = B_{o_t}...B_{o_{t+h}}$.
\end{lemma}

The proof of Lemma \ref{lemma:multi} is a simple application of the chain rule and can be found in appendix B.

We fix a horizon $H$ and average the $h$ step gradients for $h \in \{1,...,H\}$ to get the final gradient update, resulting in Algorithm \ref{alg:multi}.

\begin{algorithm}[tb]
   \caption{Learning a PSR with Multi-Step Inference Gradients}
   \label{alg:multi}
\begin{algorithmic}
   \STATE {\bfseries Input:} sequence of observations $o_{1:T}$,  corresponding set of features of future observations $\psi_{1:T}$, learning rate  $\alpha \in \mathbb{R}^+$, horizon $H \in \mathbb{N}^+$, number of iterations $n \in \mathbb{N}^+$.
   \STATE $(q_1, \{B_i\}_{i=1}^k, b_{\infty}) \leftarrow$ \textbf{2S-Regression}($o_{1:T}$, $\psi_{1:T}$)
   \FOR{$i=1$ {\bfseries to} $n$}
   \STATE $\hat{q}_1 \leftarrow q_1$
   \FOR{$t=1$ {\bfseries to} $T$}
   \STATE $\Delta \leftarrow 0$
   \FOR{$h=1$ {\bfseries to} $H$}
   \STATE $\hat{q}_{t+h} \leftarrow\frac{B_{o_{t:t+h-1}}\hat{q}_t}{b_{\infty}^T B_{o_{t:t+h-1}}\hat{q}_t}$
   \STATE $\Delta \leftarrow \Delta + \frac{\partial}{\partial A} \frac{1}{2}\norm{ \psi_{t+h} - \hat{q}_{t+h} }_2^2$
   \ENDFOR
   \STATE $\hat{q}_{t+1} \leftarrow \frac{B_{o_t}\hat{q_t}}{b_{\infty}^T B_{o_t}\hat{q}_t}$
   \STATE $B_{o_t} \leftarrow B_{o_t} - \alpha\Delta$
   \ENDFOR
   \ENDFOR
\end{algorithmic}
\end{algorithm}

Note that Multi-step inference gradients are equivalent to
Backpropagation Through Time (BPTT) for recurrent
neural networks. Indeed, the PSR update Equation \ref{eqn:psr_update} defines a special form of a recurrent structure.

\section{Evaluation Metrics}
\label{sec:metrics}
We evaluate each model using a variety of metrics in order to examine how the model performs on various inference tasks.

As discussed earlier the log likelihood is not well defined for all PSR model parameters, hence we evaluate all models on the \emph{Proxy Negative Log Likelihood (PNLL)} proposed by \cite{jks2016} (see equation \ref{eqn:pnll}). A good PSR is one which assigns high PNLL to sequences of observations in the test set. We expect the model of \cite{jks2016} to perform well on PNLL, as their model directly optimizes this function. We note that PNLL is a specific artificial metric defined by \cite{jks2016} for their model, and that while suggestive, does not necessarily imply good predictive performance.

Dynamical systems models such as PSRs are used in practice to predict future observations. These predictions can be one-step predictions (the next observation) or multi-step predictions (an observation at some point in the future). An example of this would be predicting the next character/word in a string of text. An appropriate evaluation metric for this task is the \emph{One-Step Prediction Accuracy (OSPA)}, i.e. the fraction of ($1$-step) predicted observations which match the true ($1$-step) observations. A larger OSPA is better.

Finally we are also interested in the quality of a PSR as a model. Our previous metrics have evaluated the quality of our PSR based on its ability to perform tasks for us. However at its core a PSR is model which allows us to transform states into new states, where each state corresponds to observable quantities. As mentioned in section \ref{sec:ig} we would like a model which makes good state updates on all states encountered during inference. Therefore our final evaluation metric will be the \emph{L2 state error (L2SE)} as defined in Equation \ref{eqn:loss_function}. A smaller L2SE is better.

Note that inference can result in pathological states, which cause very large values of the squared loss. In fact it was this exact behavior which motivated this work. To allow us to analyze this behavior we examine both the mean and median of the L2SE. The median provides us with a robust estimate of the general trend, while the mean provides us with insight into the number and magnitude of these pathological states.

A good model will perform well on all of these metrics. It is particularly important for our trust in the algorithm that if we optimize our model with respect to one metric that it should perform well on other metrics. Note that Inference Gradients and PSIM optimize with respect to L2SE, while Jiang et al. optimize with respect to PNLL.

\section{Experiment: Synthetic HMM}
\label{sec:ring}
In our first experiment we compare the performance of Inference Gradients, Multi-step Inference Gradients, 2-stage regression, PSIMs, and the likelihood based gradient approach of \cite{jks2016} on a synthetic HMM\@. This mirrors an experiment performed in \cite{jks2016}.

\subsection{Dataset}
We generate $10^4$ length 10 observation sequences from a randomly generated ring-topology HMM with $20$ states and $20$ observations. Each state has at most 2 possible observations, chosen at random. The transition matrix follows a ring topology, where each state can only transition to its two neighbors or to itself. All non-zero entries of the transition matrix, the emission matrix, and the initial state distribution are picked uniformly randomly from $[0, 1]$ and then normalized. We split the observation sequences equally into training and test data.

\subsection{Parameters}
We use all strings of length 1 and 2 as our features. We evaluate the performance of the learned PSRs on each of the metrics discussed in Section \ref{sec:metrics}. We average results over 100 independent trials. 

For all methods we normalize gradients to have an L1 norm of one, and use a fixed learning rate of $10^{-3}$ (selected via cross validation). We believe this is the fairest way of comparing the performance of these disparate approaches, as it means the only difference between gradient updates is the direction of the gradient step. 

\begin{figure*}[!t]
\centering
\includegraphics[scale=\proxyscalefactor]{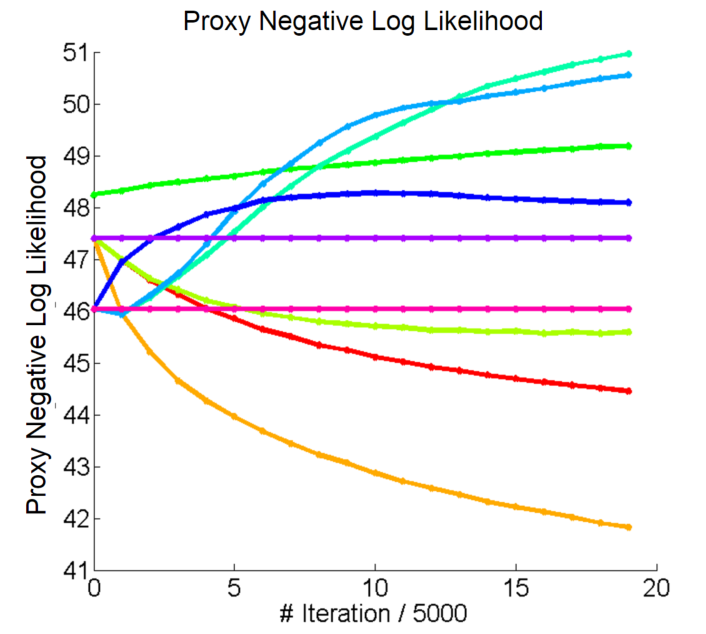}
\includegraphics[scale=\scalefactor]{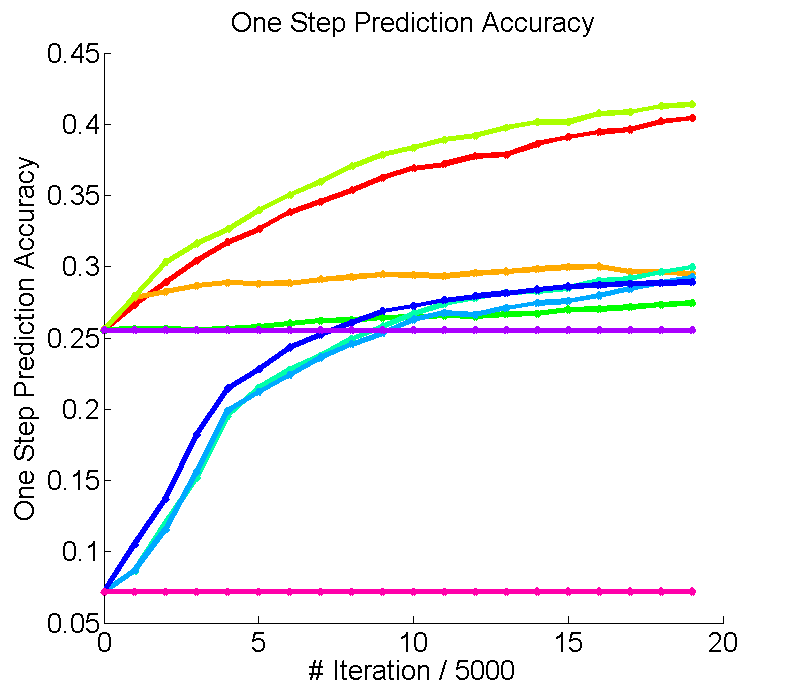}
\includegraphics[scale=\scalefactor]{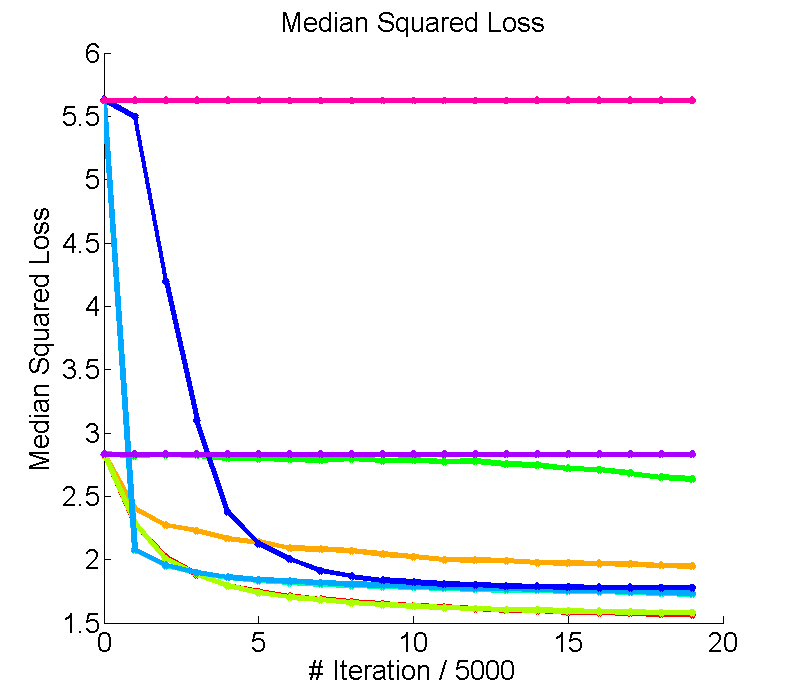}
\includegraphics[scale=\scalefactor]{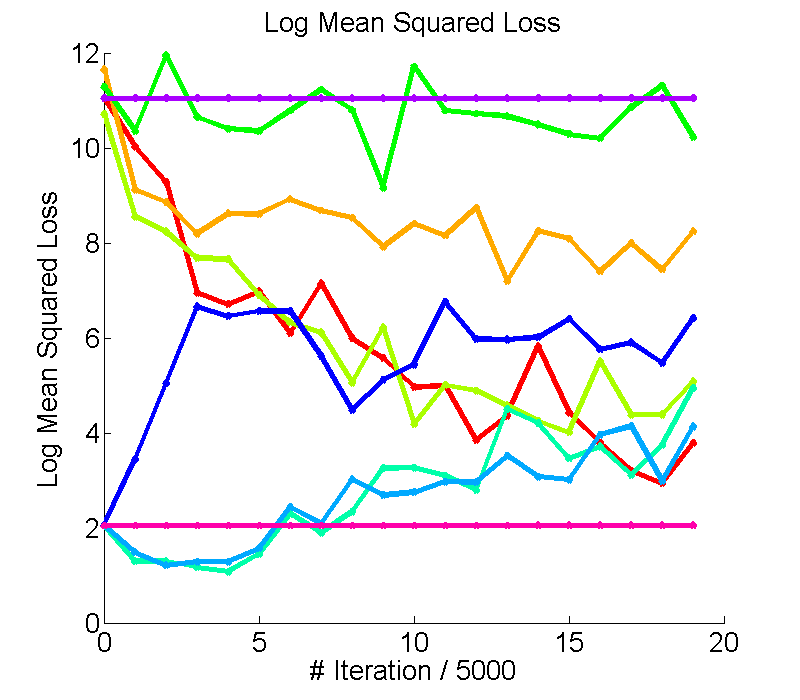}
\includegraphics[scale=\scalefactor]{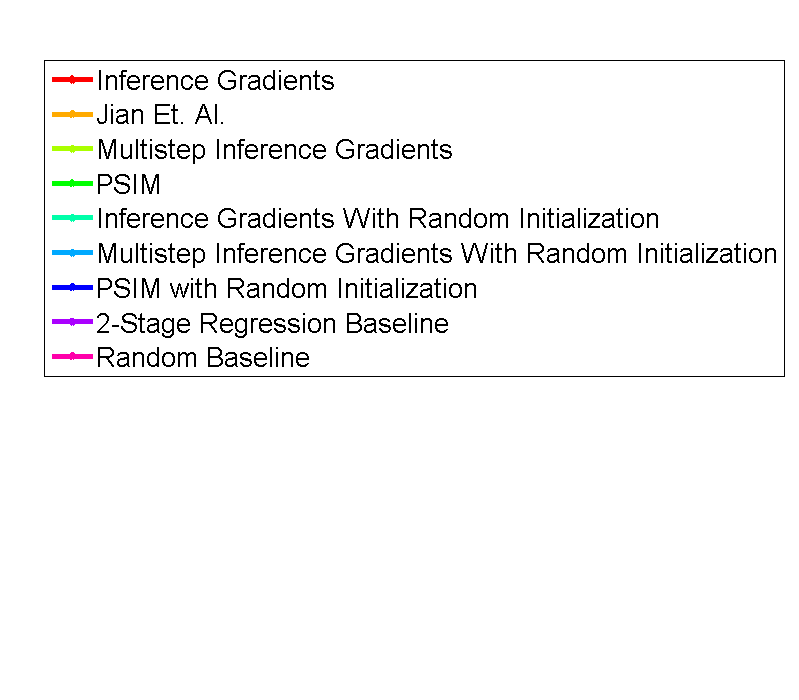}
\caption{Comparison of various approaches on the ring topology synthetic HMM dataset with respect to the number of iterations. }
\label{fig:ring}
\end{figure*}

\subsection{Discussion}

Results are presented in Figure \ref{fig:ring}. We note that the random baseline and 2SR baseline remain constant across all iterations, and hence correspond to flat lines in all figures. Additionally all results are presented in terms of the number of iterations, however in practice Inference Gradients was also several orders of magnitude faster than the approach of Jiang et al. for the same number of iterations, and had comparable speed to PSIM\@.

\begin{itemize}
    \item \textbf{Proxy Negative Log Likelihood:} Inference Gradients, Multi-step Inference Gradients, and Jiang et al. all decrease the PNLL, while all other approaches increase the PNLL\@. Jiang et al. decrease the PNLL the most, which is not surprising given that their approach performs gradient descent on a surrogate to the PNLL\@. It is important to note that Inference Gradients and Multi-step Inference Gradients both decrease the PNLL even though they are performing gradient descent on the L2SE\@. Finally it is clear that the 2SR initialization is extremely important, as both Inference Gradients and Multi-step Inference Gradients increase the negative log likelihood when initialized randomly.
    \item \textbf{One Step Prediction Accuracy:} All approaches improve the OSPA\@. Multi-step Inference Gradients results in the best OSPA, with Inference Gradients a close second. Note that Multi-step Inference Gradients is outperforming Inference Gradients on OSPA despite being optimized for Multi-step prediction. This suggests that optimizing model parameters based on long term effects is a good strategy. The first few iterations of Jiang et al. result in a small increase in OSPA, however OSPA quickly plateaus, with subsequent iterations failing to further improve OSPA, despite continuing decreases in the PNLL\@. This result highlights the importance of evaluating the learned PSR according to multiple metrics, and suggests that optimizing the PSR with respect to the negative log likelihood may not be the best approach for optimizing a PSR if the goal is to use that PSR for prediction tasks. Once again we note the importance of the 2SR initialization: the OSPA of randomly initialized Inference Gradients and randomly initialized Multi-step Inference Gradients is significantly lower. Furthermore it appears to plateau, suggesting the possibility of local optima. Even if this is not the case it will take far more iterations to achieve the same level of performance. 
    \item \textbf{Median L2 state error:} The median L2SE results mirror the OSPA results, with Multi-step Inference Gradients outperforming all other approaches. This is to be expected as Inference Gradients optimizes the model with respect to L2SE.
    \item \textbf{Log Mean L2 state error:} All approaches which were not randomly initialized improve log mean L2SE, while all randomly initialized approaches increase it. One of the original inspiration behind Inference Gradients was to use techniques from imitation learning to improve model predictions on unseen states. We theorized that improving model predictions on unseen states would result in improved performance on other metrics. \emph{This result shows that Inference Gradients is achieving exactly this behavior}. This decrease shows that Inference Gradients decreases the number and magnitude of pathological states encountered during inference. It is interesting to note that randomly initialized approaches increase log mean L2SE while decreasing median L2SE\@. In other words these methods improve most states, while producing more rogue states.
\end{itemize}

\section{Experiment: English Text}
In our second experiment we compare the performance of the same set of approaches on an English text dataset. We note that in this dataset OSPA corresponds to the accuracy when the model is used to predict the next character in a text string.  This experiment also mirrors one performed by Jiang et al. \cite{jks2016}.

\begin{figure*}[!t]
\centering
\includegraphics[scale=\proxyscalefactor]{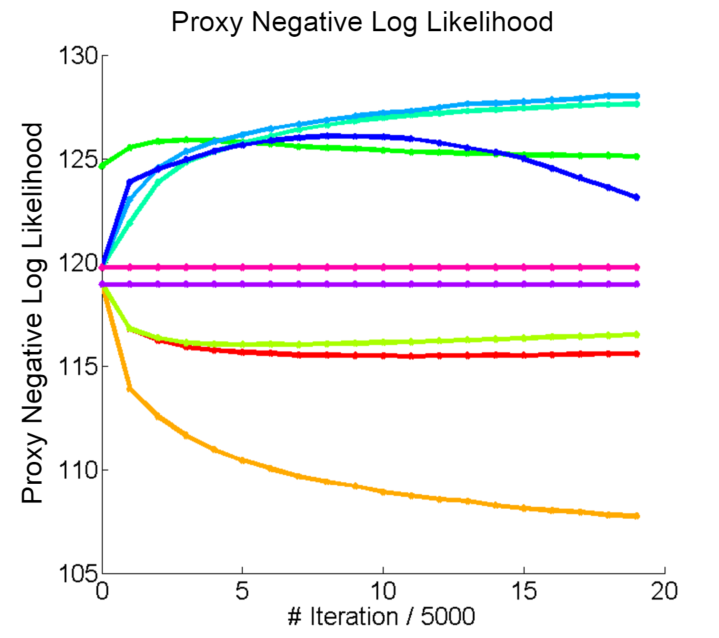}
\includegraphics[scale=\scalefactor]{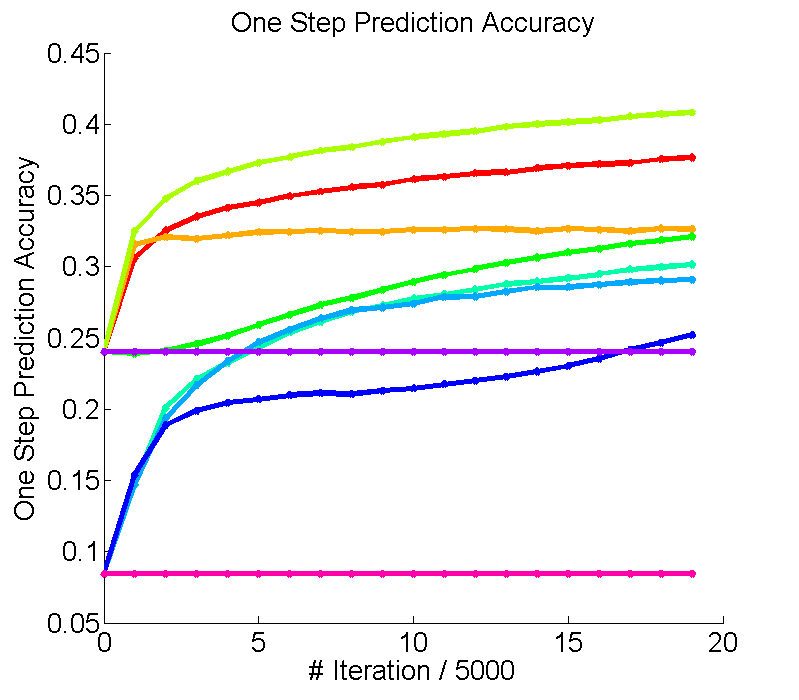}
\includegraphics[scale=\scalefactor]{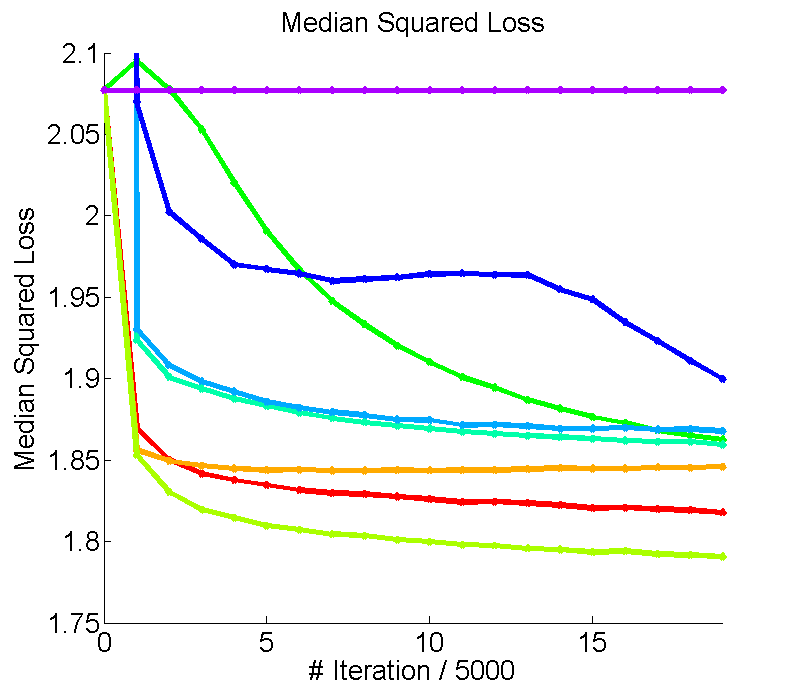}
\includegraphics[scale=\scalefactor]{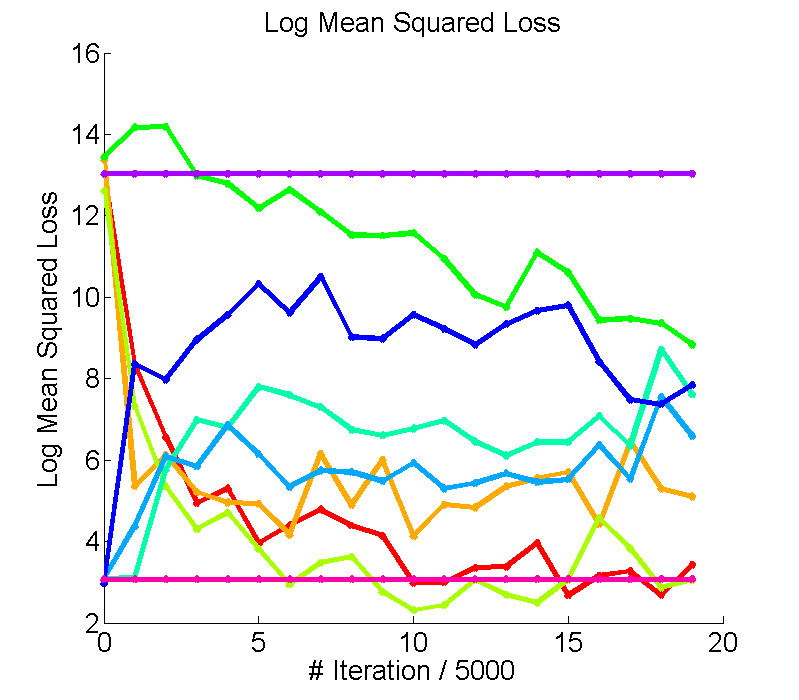}
\includegraphics[scale=\scalefactor]{legend.png}
\caption{Comparison of various approaches on an english text dataset with respect to the number of iterations. }
\label{fig:text}
\end{figure*}

\subsection{Dataset}
We sample random length $10^5$ excerpts from the Penn Tree Bank (PTB) split evenly into training and test data. 

\subsection{Parameters}
We use all strings of length 1 and 2 as our features. We evaluate the performance of the learned PSRs on each of the metrics discussed in Section \ref{sec:metrics}. We average results over 100 independent trials. For all methods we normalize gradients to have an L1 norm of one, and use a fixed learning rate of $10^{-3}$ (selected via cross validation). \footnotetext{Cross validating each method independently resulted in similar optimal learning rates for all methods, hence for simplicity we fixed the learning rate to be constant for all methods. We note that this offers the additional benefit of allowing us to compare performance based only on the direction of the gradient.} 

\subsection{Discussion}
Results are shown in Figure \ref{fig:text}. These results match those from the previous experiment. The only (minor) difference is that Multi-step Inference Gradients gives a larger performance improvement over Inference Gradients on this, more complex, dataset.

\section{Conclusions}
We presented Inference Gradients, a simple, fast, and robust method for practical learning of PSRs.

Spectral algorithms for learning PSRs are single pass algorithms. This means that they are not designed to minimize errors resulting from recursive operations such as the state updates in filtering and prediction. Therefore the PSR often makes poor predictions on the additional states encountered during inference. We present a simple algorithm which refines the PSR to make good state predictions from all states encountered during inference on test data. Our approach consists of simple gradient updates inspired by ideas from imitation learning.

This approach provides us with a method to refine the model parameters of a PSR which has been initialized using a spectral algorithm. This refinement optimizes predictive performance and minimizes the predictive error without resorting to approximations to the Likelihood. We show that our approach outperforms the current state-of-the-art gradient descent algorithm for PSRs on several metrics.


\bibliographystyle{icml2017}
\bibliography{nips_2016}

\newpage
\appendix

\onecolumn

\section{Proof of Lemma \ref{lemma:1step}}
\label{apdx:1step}

$$\frac{\partial}{\partial A} \frac{1}{2}\norm{ y - \frac{Ax}{z^TAx}}_2^2 = \left(\frac{zx^TA^T-(z^TAx) I}{(z^TAx)^2}\right)(y-\hat{y})x^T$$

\begin{proof}
\begin{align*}
    &\frac{\partial}{\partial A_{ij}} \frac{1}{2}\norm{ y - \frac{Ax}{z^TAx}}_2^2\\
    &= \frac{\partial}{\partial A_{ij}}\frac{1}{2}\sum_k \left( y_k - \frac{\sum_bA_{kb}x_b}{\sum_{kj}z_{k}A_{kj}x_j}\right)^2\\
    &= \sum_k \left(\frac{\partial}{\partial A_{ij}}\frac{-\sum_bA_{kb}x_b}{\sum_{ab}z_aA_{ab}x_b}\right)\left(y_k - \frac{\sum_bA_{kb}x_b}{\sum_{ab}z_aA_{ab}x_b}\right)\\
    &=\sum_{k\neq i} \left( \frac{\partial}{\partial A_{ij}}\frac{-\sum_bA_{kb}x_b}{\sum_{ab}z_aA_{ab}x_b}\right)\left(y_k - \frac{\sum_bA_{kb}x_b}{\sum_{ab}z_aA_{ab}x_b}\right) + \left( \frac{\partial}{\partial A_{ij}}\frac{-\sum_bA_{ib}x_b}{\sum_{ab}z_aA_{ab}x_b}\right)\left(y_i - \frac{\sum_bA_{ib}x_b}{\sum_{ab}z_aA_{ab}x_b}\right)\\
    &=\sum_{k\neq i} \left( \frac{z_i x_j\sum_bA_{kb}x_b}{(z^TAx)^2}\right)\left(y_k - \frac{\sum_bA_{kb}x_b}{z^TAx}\right) + \left( \frac{-x_j z^TAx + z_i x_j\sum_bA_{ib}x_b}{(z^TAx)^2}\right)\left(y_i - \frac{\sum_bA_{ib}x_b}{z^TAx}\right)\\
    &=\sum_k \left( \frac{z_i x_j\sum_bA_{kb}x_b}{(z^TAx)^2}\right)\left(y_k - \frac{\sum_bA_{kb}x_b}{z^TAx}\right) + \left( \frac{-x_j z^TAx}{(z^TAx)^2}\right)\left(y_i - \frac{\sum_bA_{ib}x_b}{z^TAx}\right)\\
    &= z_i\left( \frac{\sum_bx_b\sum_kA_{kb}(y_k - \hat{y}_k)}{(z^TAx)^2}\right)x_j - \left( \frac{1}{z^TAx}\right)(y_i - \hat{y}_i)x_j\\
    &= z_i\left( \frac{(y-\hat{y})^TAx}{(z^TAx)^2}\right)x_j - \left( \frac{1}{z^TAx}\right)(y_i - \hat{y}_i)x_j\\
    &= \left(\left(\frac{zx^TA^T-(z^TAx) I}{(z^TAx)^2}\right)(y-\hat{y})x^T\right)_{ij}
\end{align*}
\end{proof}

\newpage
\section{Proof of Lemma \ref{lemma:multi}}
\label{apdx:multi}

\begin{align*}
    &\frac{\partial}{\partial A} \frac{1}{2}\norm{ y - \frac{A\frac{Bx}{z^TBx}}{z^TA\frac{Bx}{z^TBx}}}_2^2
    = \frac{\partial}{\partial A} \frac{1}{2}\norm{ y - \frac{ABx}{z^TABx}}_2^2\\
\end{align*}

\begin{proof}
\begin{align*}
    &\frac{\partial}{\partial A_{ij}} \frac{1}{2}\norm{ y - \frac{ABx}{z^TABx}}_2^2\\
    &= \frac{\partial}{\partial A_{ij}}\frac{1}{2}\sum_k \left( y_k - \frac{\sum_{ab}B_{ka}A_{ab}x_b}{\sum_{abc}z_{a}B_{ab}A_{bc}x_c}\right)^2\\
    &= \sum_k \left(\frac{\partial}{\partial A_{ij}}\frac{-\sum_{ab}B_{ka}A_{ab}x_b}{\sum_{abc}z_{a}B_{ab}A_{bc}x_c}\right)\left( y_k - \frac{\sum_{ab}B_{ka}A_{ab}x_b}{\sum_{abc}z_{a}B_{ab}A_{bc}x_c}\right)\\
    &=\sum_{k} \left(\frac{-B_{ki}x_j\left(\sum_{abc}z_{a}B_{ab}A_{bc}x_c\right) + \sum_{ab}B_{ka}A_{ab}x_b\left(\sum_{a}z_{a}B_{ai}x_j\right)}{\left(\sum_{abc}z_{a}B_{ab}A_{bc}x_c\right)^2}\right)
    \left( y_k - \frac{\sum_{ab}B_{ka}A_{ab}x_b}{\sum_{abc}z_{a}B_{ab}A_{bc}x_c}\right)\\
    &=\sum_{k} \left(\frac{-B_{ki}x_j(z^TBAx) + [BAx]_k\left([zB]_i  x_j\right)}{(z^TBAx)^2}\right)
    \left( y_k - \left[\frac{BAx}{(z^TBAx)}\right]_k \right)\\
    &=\left(\frac{[(y-\hat{y})^TB]_ix_j(z^TBAx) + \left((y-\hat{y})^TBAx\right)\left([zB]_i  x_j\right)}{(z^TBAx)^2}\right)\\
    &=\left(\frac{B^Tzx^TA^TB^T- B^T(z^TBAx)}{(z^TBAx)^2}\right)(y-\hat{y})x^T\\
\end{align*}
\end{proof}

\end{document}